\documentclass[conference]{IEEEtran}

\ifCLASSINFOpdf
  \usepackage[pdftex]{graphicx}
  \graphicspath{ {pictures/}{../pdf/}{../jpeg/} }
  \DeclareGraphicsExtensions{.pdf,.jpeg,.png,.PNG}
\else

\fi

\ifCLASSOPTIONcompsoc
 \usepackage[caption=false,font=normalsize,labelfont=sf,textfont=sf]{subfig}
\else
 \usepackage[caption=false,font=footnotesize]{subfig}
\fi

\usepackage{flushend}
\usepackage{tabularx,booktabs,textcomp}
\newcolumntype{C}{>{\centering\arraybackslash}X} 
\newcolumntype{L}{>{\raggedright\arraybackslash}X} 
\usepackage[noadjust]{cite}

\usepackage{wasysym}
\usepackage{lipsum} 
\usepackage[table]{xcolor}
\usepackage{adjustbox}
\usepackage{graphicx}
\usepackage{dirtytalk}
\usepackage{enumitem}
\usepackage{colortbl}
\usepackage{hhline}
\usepackage{multirow}

\begin{document}

\title{A Review on Trust in Human-Robot Interaction}

\author{\IEEEauthorblockN{Zahra Rezaei Khavas}
\IEEEauthorblockA{Department of Electrical and Computer Engineering\\
University of Massachusetts Lowell\\
Lowell, Massachusetts\\
Email: Zahra\_RezaeiKhavas@student.uml.edu}}

\maketitle

\thispagestyle{plain}
\pagestyle{plain}


\begin{abstract}
\bstctlcite{IEEEexample:BSTcontrol}
Due to agile advancements in the autonomy of robotic systems, prospective robotic agents are intended to play the role of teammates and partners with humans to perform operations rather than tools that are replacing humans or helping humans in specific tasks. This notion of “partnering” with robots raises new challenges for the human-robot interaction (HRI), which gives rise to a new field of research in HRI, namely Human-Robot Trust. Where humans and robots are working as partners, the performance of work can be diminished if humans do not trust robots appropriately. Considering the impact of human-robot trust observed in different HRI fields, many researchers have investigated the field of human-robot trust and examined various concerns related to the human-robot trust. In this work, we review the past works on human-robot trust based on the research topics and discuss selected trends in this field. Based on these reviews, we finally propose some ideas and areas of potential future research at the end of this paper.
\end{abstract}



\section{Introduction}

Trust is one of the essential elements in the development of any constructive relationship. Trust is not limited to only inter-personal interactions, and it can affect different forms of interaction, including relationships among human individuals and robots. We can mention trust as an overarching concern that affects the effectiveness of a system, especially in terms of safety, performance, and use rate \cite {lee2004trust}. Having this concern in mind, trust has become a critical element in the design and development of automated systems \cite{martelaro2016tell}. Autonomous systems are being designed and developed with increased levels of independence and decision-making capabilities, and these capacities will be efficient in uncertain situations \cite{schaefer2016meta}.

Human-robot trust is an important branch of human-robot interaction (HRI) which has recently gained increasing attention among scholars in many disciplines, such as Computer Engineering \cite{robinette2016overtrust}, Psychology \cite{hancock2011meta}, Computer Science \cite{desai2012modeling} and Mechanical Engineering \cite{sadrfaridpour2016modeling}. Trust is a significant factor that needs to be taken into considerations when robots are going to work as teammates in human-robot teams \cite{groom2007can}, where robots are designed to be used as autonomous agents \cite{selkowitz2015effects}, or where robots are going to be used in a complex and dangerous situation and take the place of the human in the high-risk tasks \cite{ososky2014determinants,robinette2016overtrust}. In many cases, trust is the main factor determining how much a robotic agent would be accepted and used by the human \cite {parasuraman1997humans}. A human would not use or rely on an automated system if they believe that system is not trustworthy \cite{lee2008review}.

Robotic is becoming more prevalent, and robots are used in variety of domains such as military \cite{voth2004new,springer2013military}, medicine \cite{taylor2006perspective,taylor2016medical}, and airspace \cite{yim2003modular,moosavian2007free}, and consumer \cite{ivanov2017robot,pandey2016socially} etc. Poor interaction between human operators and robots will become progressively costly and calamitous. Week partnerships as the result of inappropriate trust between humans and robotic agents might cause misuse and disuse of a robotic agent. Misuse refers to the failures that take place as the result of the user’s over-trust in the robotic agent and accepting all the solutions and results presented by the robotic agent without questioning. Disuse refers to the failures that occur as the result of human under-trust in the robotic agent and rejecting the capabilities of robotic agent \cite{lee2004trust}. In order to prevent human operators’ misuse and disuse of robots, different aspects of trust such as the definition of trust and factors affecting trust in HRI needs to be investigated thoroughly to generate trust measurement methods, trust models, and trust repair strategies to calibrate trust \cite{ososky2013building}. 

In this work, we review the past works on human-robot trust based on the research topics and discuss selected trends in this field. Based on these reviews, we finally propose some ideas and areas of potential future research at the end of this work. The overall purpose of this document is to explore different studies concentrated on trust in HRI and review these studies. The selected trends of human-robot trust that are going to be discussed in this work are as follow: first, we review different definitions of trust and talk about the multidimensional nature of trust in HRI, then we will talk about factors affecting trust in HRI and classify those factors, after that we will talk about trust repair and trust calibration and finally, we will talk about modeling trust and trust measurement techniques. Based on the reviews and comparison of the available works on trust in HRI, shortcoming and challenges of studying trust in HRI that has not been answered yet and more avenues of research need to be concentrated on them, are mentioned in the conclusion and future work section. 

\section{Definition of Trust in HRI}
From the psychology point of view, trust is a mental state of a human \cite{atoyan2006trust}. Numerous researchers have extensively explored the notion of trust for decades. Trust is not limited to just interpersonal interaction. It underlies different forms of interaction, such as banks' interactions with customers, governments with citizens, employers with employees, etc., \cite{wagner2011recognizing}. Therefore, we can say trust can affect human-robot interaction, as it can affect the human user or collaborator's willingness to assign a task, share information, cooperate with, provide support, accept results, and interact with a robot \cite{freedy2007measurement}.

As trust is one of the necessities for building a successful human-robot interaction, we need to study trust and factors affecting the formation and loss of trust to create methodologies for modeling, measuring, and calibrating trust. The first step toward studying trust is having a clear definition of trust. However, despite the broad efforts and a considerable number of studies concentrating on trust, yet there is no unique definition for trust, as the definition of trust concept is heavily dependent on the context in which trust is being discussed \cite{cameron2015framing}. There are many different definitions for trust by different researchers in human-robot trust field, for instance, in \cite {oleson2011antecedents} trust is defined as \textit{“the reliance by one agent that actions prejudicial to the well-being of that agent will not be undertaken by influential others”}.

The issue for making a unique definition for trust within the context of HRI comes primarily from a variety of applications, and situations that robots are used, around which trust needs to be defined, explored, and measured as a variable. Different applications, different robotic agents, different human operators, and different operation situations would provide different challenges for defining trust for every specific context in HRI that robots are used. For example, in robotic applications that trust needs to be explored in the context of user-safety and high-risk situation such as emergency evacuation \cite{robinette2016overtrust}, the definition of trust might differ substantially from the situation in which trust is explored in the context of border tracking robots \cite{xu2015optimo}.

One of the most thorough definitions of trust, which is deployed by many other studies concentrated on human-robot trust, is by Lee and See \cite{lee2004trust}. They define trust from the perspective of automation. This definition was generated by reviewing many other studies concentrated on defining trust and was complementary to many other works. They define trust as \say{\textit{the attitude that an agent will help achieve an individual's goals in a situation characterized by uncertainty and vulnerability}}. This definition of trust is accepted and used by many studies on trust in HRI. Wagner et al. \cite{wagner2011recognizing} also provided a comprehensive definition for trust: \say{\textit{a belief, held by the trustor, that the trustee will act in a manner that mitigates the trustor's risk in a situation in which the trustor has put its outcomes at risk}}. They also provided a model for determining if an interaction demands trust or not. All these definitions have two things in common, one is:``whether robot's actions and behaviors correspond to human's interest or not?", and two is: "human has something to lose." To address this concern in each robotic application, trust needs to be defined based on human interest in that domain.

\section{Multidimensional Nature of Trust in HRI}

In recent years due to the advancements of robotics, robots are being used very close contact with humans in a variety of settings, ranging from homes \cite{frennert2017case} and schools \cite{ahtinen2020learning}  to the workplace \cite{reeder2010breakbot}. Robots are no more used as tools that their only contribution is to perform repetitive physical tasks. With the advances in the realm of robotic and human-robot interaction, robots are increasingly used as social agents in a variety of social applications \cite {wolbring2014social}. Therefore, to investigate the nature of trust in human-robot interaction, one should move from the domain of human-automation trust into the domain of human-human trust \cite{malle2021multidimensional}.

Works in the domain of human-automation trust emphasize the performance of automated systems. In the literature on trust in automation, the main focus is on improving human users' trust in automation by modifying the performance of the system based on human expectation or matching a human user's expectations of a system with information about the system performance. Automated systems are motiveless, and there is no mention of any concerns about being deceived, betrayed, or exploited by the system. In sum, trust in automated systems is highly correlated with the system performance, and the only worries concern the systems' reliability and ability \cite{schaefer2016meta}. However, the emphasis of the works in the domain of human-human trust is on moral trust. The question is whether a human agent will take advantage of another human's vulnerability either unconsciously because of a lack of ability or consciously because of a lack of moral integrity \cite{park2020multifaceted}.

A modern conceptualization of human-robot trust which is not limited to the traditional interpretation of human-automation trust and better matches the current understanding of human-robot trust, is provided by Ulman et al. \cite{malle2021multidimensional}. They reviewed the literature and performed empirical work, finally they suggested that trust is multidimensional, incorporating both performance aspects (i.e., central in the human-automation literature) and moral aspects (i.e., central in the human-human trust literature). These two trust aspects break into two sub-factors (i.e., Reliable and Capable within Performance, and Sincere and Ethical within Moral). A multidimensional conceptualization of trust in HRI can be applied to any robotic applications which demands trust in robots. However, only some of the trust dimensions may be relevant for any given robotic application or interaction type. 

Based on the multidimensional nature of the trust in human-robot interaction, Park et al. \cite{park2020multifaceted} divide the human-robot trust into two categories, performance-based trust, and relation-based trust. Performance-based trust mainly emphasis on reliability, capability, and competency of the robot at any given task, without demanding to be monitored by a human supervisor. A robot that garnered the performance-based trust of a human’s trust is a proper choice to perform a critical factory job with zero interaction with humans. However, relation-based trust implies the acceptance of a robot as a trusted social agent. A robot should have some features such as being sincere and ethical to the person with whom it interacts. A robot that is granted relation-based trust is a good choice for robotic uses that involve close interaction with a human. This type of trust is less studied in the past, but it is becoming increasingly relevant by the advancements in social robots, and therefore more researches will be concentrated on factors that may influence this type of trust.

\section{Factors Affecting Trust}

Most of the current researches in the human-robot trust are concentrated on factors affecting trust (e.g., trust gain, trust loss and trust repair in human after robot's faulty behavior or trust violation). The top factors affecting human-robot trust can be considered an extension to the factors affecting human-automation trust. Lee et al. \cite{lee2004trust} provided a complete review of the factors affecting trust in human-human interaction and generalize these factors by theoretical and empirical examination to factors affecting trust in human-automation interaction (HAI). There are also many other studies reviewing and analyzing factors affecting trust in HAI \cite{hoff2015trust,schaefer2016meta}. However, robots differ from other forms of automation in many cases. First, robots are mobile, and this mobility can cause injury to the human when they are working in the same field. Second, robots can work remotely, and they can go out of control. Third, robots are designed in many different shapes, which sometimes people are not familiar with those, and it affects the whole trust process as well \cite{desai2012modeling}. Therefore, although there are many similarities among factors affecting trust in human-automation and human-robot interaction, factors affecting trust in human-robot interaction need to be investigated separately. 

To identify the elements that affect trust between humans and robots, researchers design experiments in which humans interact with robots in different scenarios. Researchers manipulate factors in the experiment that want to examine their effect on trust. There are numerous studies in the field of HRI that study the effects of different factors on trust. Hancock et al. \cite{hancock2011meta} provided a meta-analysis of factors affecting trust in human-robot interaction and classify these factors in three classes: Human-related factors (i.e., ability-based, characteristics); Robot-related factors (i.e., performance-based, attribute-based); and Environment-related factors (i.e., team collaboration, tasking). In this study, we will review and classify recent studies on modeling trust in human-robot interaction and factors affecting trust with a similar classification basis as \cite{hancock2011meta} with some changes in categories and subcategories. In this work, factors affecting trust in human-robot interaction are classified into three categories: 1- Robot-related factors (i.e., robot-performance, robot’s appearance, robot’s behaviors, and physical presence of the robot), 2- task and environment-related factors, 3- human-related factors. Table \ref{tab:my-table} shows our classification of factors affecting trust.

\begin{enumerate}
\item\textbf{Robot-Related Factors:}
These factors are directly related to the design and construction of the robots, and by changing the design and construction of the robot, trust in robots can be increase or decrease. Robot-related factors have the most significant effect on the trust in human-robot interaction \cite{hancock2011meta}. This justifies the significant number of researches in the human-robot trust field, concentrating on robot-related factors. In this study, we classify robot-related factors under four subcategories: performance-related factors, behavior-related factors, appearance-related factors, and a single feature: robot physical presence.

\begin{enumerate}[label=(\alph*)]
\item{Performance-Related Factors:}
There are many empirical and analytical studies over performance-related factors affecting trust in human-robot interaction. Performance-related factors determine the quality of the performed operation by the robot from the human operator's point of view. Of these factors we can mention reliability, faulty behavior, frequency of fault occurrence \cite{desai2012effects}, timing of error \cite{lucas2018getting}, transparency, feedback \cite{salem2015would,natarajan2020effects}, level of situation awareness \cite{boyce2015effects}, false alarms \cite{yang2017evaluating}, and level of autonomy \cite{lazanyi2017dispositional}.

\item{Appearance and Personality-Related Factors:}
People consider a personality for robots based on robots' appearance and behavior while interacting with them. Some features in robot appearance such as anthropomorphism \cite{natarajan2020effects}, robot's gender \cite{tay2014stereotypes}, harmony of robot's activity with it's appearance\cite{robinette2013building}, and robot's similarity (e.g., the same/different gender) by two deep level human-robot similarity (e.g., similar/different work style)\cite{you2018human} affect trust in human-robot interaction.
\item{Behavior-Related Factors:}
 Advancements in robotic systems in recent years and increased autonomy of robots caused people to consider them more like teammates to tools, and increase the intelligence of robots altered the form of human-robot interacting to a more naturalistic interaction\cite{ososky2013building}. Approaching a more human-like interaction with robots cause people to consider the intention for robot's behavior. Some of the robot's behaviour such as robot's likeability (e.g., gaze behaviors and greeting)\cite{mumm2011human}, proximity (e.g., physical and physiological proximity) \cite{mumm2011human,obaid2016stop,walters2011long}, engagement \cite{robinette2013building}, confess to the reliability \cite{wang2016trust,natarajan2020effects} and harmony of robot personality with the task (e.g., introverted robotic security guard and an extroverted robotic nurse) \cite{tay2014stereotypes} can affect formation and maintenance of trust. there are some other behaviors such as apologies after failure, making excuses and explanations after failure or dialogues \cite{lucas2018getting,strohkorb2018ripple,sebo2019don,natarajan2020effects} can affect trust repair after failure. Some studies on the effect of robot's behavior on trust also showed that the harmony of behaviors with robot's activity(i.e., introverted robotic security guard and an extroverted robotic nurse) could affect humans' trust toward robotic agent\cite{tay2014stereotypes}.
\item{Robot's Physical Presence:}
Robot's co-location with participants in an experiment causes people to trust the robot better than when the robot is virtually present (i.e., telepresence throw video display) in the experiment. Studies on the effects of robot physical presence on trust showed that people are more willing to trust robots and comply with their commands and even irrational commands when robots are physically present \cite{bainbridge2011benefits,natarajan2020effects}.

\end{enumerate}

\item\textbf{Task-Related Factors:}
Based on \cite{hancock2011meta} Second greatest affect on the human-robot trust belongs to task-related factors based on \cite{hancock2011meta}. Many factors related to the nature of the task such as rationality and revocability of the tasks\cite{salem2015would}, risk and human-safety \cite{robinette2016overtrust}, work-load \cite{desai2012effects}, nature of the task (i.e., operating as a nurse or soldier) \cite{tay2014stereotypes}, task duration \cite{walters2011long,yang2017evaluating} and task site \cite{lazanyi2017dispositional} are thoroughly investigated in by researchers in HRI field.

\item\textbf{Human-Related Factors:}
Although human-related factors have the least effect on trust on factors affecting trust \cite{hancock2011meta}, they are important on formation of trust and fluctuation of trust during human-robot interaction. There are many studies concentrated on investigation of the effect of human-related factors on trust. Of these factors, individual's gender and subjective feeling toward robots\cite{obaid2016stop}, initial expectations of people toward automation \cite{yang2017evaluating}, previous experience of the individuals with robots \cite{walters2011long} and also human's understating of the system \cite{ososky2013building} are some of human-related factors that are addressed in different studies.

\begin{table*}
\caption{Factors affecting development of trust in human-robot interaction}
\label{tab:my-table}
\resizebox{\textwidth}{!}{%
\centering

\begin{tabular}{|l|l|l|}
\hline
\rowcolor[HTML]{9B9B9B} 
\multicolumn{2}{|c|}{\cellcolor[HTML]{9B9B9B}1. Robot-Related}                                                                                                         & 3. Task \& Environment-Related                                                                                        \\ \hline
\multicolumn{1}{|c|}{\cellcolor[HTML]{C0C0C0}(a) Performance-Related}                     & \cellcolor[HTML]{C0C0C0}(c) Appearance-Related                             & Nature of task \cite{tay2014stereotypes,lazanyi2017dispositional}                                                                                                       \\ \hline
\begin{tabular}[c]{@{}l@{}}Dependability, reliability and \\  error \cite{desai2012effects,desai2013impact,salem2015would,robinette2016overtrust,natarajan2020effects,lucas2018getting} \end{tabular}        & Similarity with operator  \cite{you2018human}                                                  & \begin{tabular}[c]{@{}l@{}}Physical presence of robot\\ in task site \cite{bainbridge2011benefits,natarajan2020effects}\end{tabular}                                     \\ \hline
Autonomy level \cite{lazanyi2017dispositional,goodrich2001experiments}                                                                        & Gender \cite{you2018human,tay2014stereotypes}                                                                    & In-group membership    \cite{rau2009effects,groom2007can}                                                                                               \\ \hline
\begin{tabular}[c]{@{}l@{}}Situation awareness, feedback \\ and Transparency \cite{desai2013impact,natarajan2020effects,boyce2015effects,wang2016trust}\end{tabular} & \begin{tabular}[c]{@{}l@{}}Harmony of appearance with \\ task \cite{robinette2013building}\end{tabular} & Duration of interaction \cite{yang2017evaluating,walters2011long}                                                                                              \\ \hline
\multicolumn{1}{|c|}{\cellcolor[HTML]{C0C0C0}(b) Behavior-Related}                        & Anthropomorphism  \cite{natarajan2020effects}                                                          & Task site \cite{lazanyi2017dispositional}                                                                                                            \\ \hline
Dialogues \cite{lucas2018getting}                                                                                & \multicolumn{1}{c|}{\cellcolor[HTML]{9B9B9B}2. Human-Related}              & Revocability \cite{salem2015would}                                                                                                         \\ \hline
\cellcolor[HTML]{FFFFFF}Proximity \cite{mumm2011human,obaid2016stop,walters2011long}                                                        & \cellcolor[HTML]{FFFFFF}Personality \cite{salem2015would}                                       & Rationality \cite{salem2015would,robinette2016overtrust,bainbridge2011benefits}                                                                                                          \\ \hline
Likeability and friendliness \cite{mumm2011human}                                                             & Culture \cite{li2010cross}                                                                   & Risk \cite{robinette2016overtrust,lazanyi2017dispositional,you2018human}                                                                                                                 \\ \hline
\begin{tabular}[c]{@{}l@{}}Personality (harmony with\\  task) \cite{tay2014stereotypes}\end{tabular}                & \begin{tabular}[c]{@{}l@{}}Understanding of the\\  system \cite{ososky2013building}\end{tabular}     &                                                                                                                       \\ \cline{1-2}
Confess to reliability \cite{wang2016trust,natarajan2020effects}                                                                   & Demographics \cite{obaid2016stop,m2011measuring}                                                               & \multirow{-2}{*}{\begin{tabular}[c]{@{}l@{}}Workload, complexity and \\required level of multi tasking  \cite{desai2012effects}\end{tabular}} \\ \hline
Apology for failure \cite{strohkorb2018ripple,sebo2019don,natarajan2020effects}                                                                      & Subjective feeling \cite{obaid2016stop,yang2017evaluating}                                                        &                                                                                                                       \\ \hline
Engagement  \cite{robinette2013building}                                                                               & Experience with robots \cite{m2011measuring}                                                    &                                                                                                                       \\ \hline
\end{tabular}

}

\end{table*}

\end{enumerate}

\section{Trust Measurement in Human-Robot Interaction}

As mentioned in the previous section, researchers design experiments in which humans interact with robots in different scenarios to identify the elements that affect trust between humans and robots. Researchers manipulate factors in the experiment that want to examine their effect on trust. Latter, they need to assess experiment participants' gain or loss of trust in the presence of those manipulated factors, so they need to apply the trust measurement technique to evaluate trust at the end of these experiments. In human-robot trust research, there are two main methods for measuring trust, subjective trust measurement and objective trust measurement. In the rest of this section, we will review these methods, their benefits and limitations, different strategies for deployment of these methods in various studies, and examples of studies that have used these methods for measuring trust in human-robot interaction.

\subsubsection{Subjective Trust Measurement} 

The first and most dominant method for trust measurement in HRI is subjective trust measurement. This trust measurement technique involves assessing experiment participants' answers to questionnaires designed to gauge a people's trust in automated agents or specifically to the robots \cite{flook2019impact}. The main advantage of subjective trust measurement methods is the ease of use, as in these methods, information is derived from the source directly. However, there are several potential issues with these trust measurement methods; those two are noted here. First, participants of the experiments may consciously or unconsciously align their answers to the demands of the experimenters in the study. Second, peoples' states of trust attribution to a robot in a specific situation may differ heavily from the actual behavioral responses of those people to a robot in risky situations. Individuals report more attributed trust to a robot in impractical situations than they actually would do in real-world scenarios where there is vulnerability and risk \cite{baker2018toward}.

Although by the use of subjective trust measurement methods, trust can be assessed at various times during a study, trust measurement in HRI is mainly dominated by post-hoc questionnaires. This makes the subjective trust measurement methods more prone to error. They only allow for measurement of trust at a singular moment, which contradicts with varying nature of trust and limits the understanding of the dynamics of trust-building \cite{baker2018toward}.

There are some trust scales or questionnaires for trust assessment developed and validated by experts to be used in HRI research. Of those we can mention, Human-Robot Trust Scale \cite{schaefer2016measuring}, Trust in Automation Scale \cite{jian2000foundations}, Human-Robot Interaction Trust Scale \cite{yagoda2012you}, and Human-Computer Trust Instrument \cite{madsen2000measuring}. Many of the studies in HRI use these validated questionnaires. However, some studies create non-validated ad-hoc questionnaires for a single research study. The validation qualification is very important and shows the applicability of those methods in other research and the comparability of the research results with other research results \cite{schaefer2016measuring}. 
\subsubsection{Objective trust measurement} These trust measurement methods are based on analyzing how experiment participants interact with the robots, rather than relying on participants' speculation about themselves. The main advantage of these methods is that they are not prone to errors related to biased answers by participants and variation between stated trust and behavioral trust in real-world scenarios \cite{flook2019impact}. However, these methods also have some drawbacks. The first and most important drawback of objective trust measurement methods is that a researcher who uses this method for measuring trust should operationalize behaviors of people and decides that certain behaviors represent trust or distrust. Therefore, when using objective trust measurement, researchers must demonstrate people's behavior as a reflection of their trust attitude toward the robot \cite{baker2018toward}. However, a person's behaviors might sometimes vary from their attitudes about trust. From the teamwork view, people's attitudes, behaviors, and cognitions are known to be distinct from each other \cite{cannon2011team}.

Objective trust measurement methods are used less frequently in human-robot trust studies than subjective trust measurement methods. There are four different categories of objective trust measurement methods \cite{law2021trust}. We will explain these four categories below:
\begin{enumerate}

\item Task Intervention: For trust measurement in this method, researchers assign a robot to perform a task normally done by a human and ask participants to interact with the robot during this task. Trust measurement is performed by monitoring the number of times the human participant intervenes in the robot's task by changing the robot's working mood from autonomous to control mood or the number of times the human participant preventing the robot from doing the task and start doing the task by themselves. Of the researches that used this trust measurement method, we can mention \cite{pedersen2018simulations,chen2018planning}.

\item Task delegation: This objective trust measurement method is based on assigning a robot to perform a task that is normally done by a human. However, in this method, at the end of the task, participants are asked to chose to assign one robot among multiple robots, or to choose from a robot or a human to perform a second task. This objective trust measurement method is also used by many researchers in HRI field \cite{xie2019robot,rossi2018impact}.

\item Behavioral Change: This objective trust measurement method is based on observing or measuring participants' behavior when they are naturally interacting with different robots. In this trust measurement method, trust is measured differently based on the nature of the interaction for each different study \cite{jayaraman2018trust,weigelin2018trust}.

\item Following advice: In this objective trust measurement paradigm, experiment participants are assigned to perform a task with the help of a robot, the robot can only give suggestions or offer advice to the participants, and participants have the option of whether or not to follow the robot advice. Trust is measured by monitoring how much participants conform with the robot's advice, and suggestion \cite{christensen2019reducing,gombolay2018robotic}.

\end{enumerate}

\section{Trust Violation and Trust Repair}

Even the robots with the most precise design will sometimes fail or make an error while performing their tasks. Researches have shown that even one mistake from the robot's side can cause a loss of trust by human users \cite{robinette2015timing}. Many researchers have investigated the effect of faulty behavior of robots on human trust \cite{salem2015would,desai2013impact}. Some researchers also investigated the effect of some special features of an error on trust such as error type \cite{flook2019impact}, timing and magnitude of the error \cite{rossi2017timing}, frequency of the error occurrence \cite{desai2012effects}, and revocability of the error \cite{salem2015would}. There are also researches on the effect of the error on trust in some special circumstances such as high-risk time-critical situations \cite{rossi2017timing,robinette2016overtrust}. When an error from the robot side occurs, possible negative effects of that error on human trust need to be mitigated. This is where trust repair comes into play. Trust repair in HRI can be defined as efforts for rebuilding trust when a robotic agent violates the trust of the human user or cooperator. But what causes the occurrence of trust violations, and how can trust be repaired after trust violation? Not enough research has been done in the area of trust repair after trust violation in the field of HRI \cite{baker2018toward}.

From the engineering perspective, robots are machines and do not intend their behaviors, and they behave the way they are designed to do. However, most of the studies on trust repair in human-robot interaction use the same strategies as those of inter-human trust repair. Some of the strategies used in human-robot trust repair are apologies, promises, compensation, showing vulnerable behavior, making excuses or explanations and dialogues, showing awareness of its error and expression of regret \cite{robinette2015timing,lee2010receptionist,hamacher2016believing,sebo2019don,strohkorb2018ripple,lucas2018getting}. The results of these studies show that taking robots as intentional agents from the psychological perspective can make users attribute sufficient beneficence to the robot's motives and can help trust repair \cite{tolmeijer2020taxonomy}. Robinette et al. \cite{robinette2015timing} also investigated that trust repair efforts such as promising to perform better, apologizing for mistakes, and convincing a human user to trust a robot by providing additional information after a failure can work, if the timing for these actions is right.

Tolmeijer et al. \cite{tolmeijer2020taxonomy} provided a taxonomy of different error or failure types that can occur during human-robot interaction and influence human users' trust toward the robot. Four error types that are mentioned in this work are 1- system error, 2- design error, 3- expectation error, and 4- user error. In this work, some mitigation strategies based on the error type are also proposed, namely fixing the error, and redesigning interaction in case of system error or designing error, providing an explanation, and providing training in case of user and expectation error, etc.

\section{Trust Modeling}

Inappropriate trust can be problematic in any form of human-robot collaboration. Both under-trust and over-trust to a robotic system might be problematic and costly \cite{lee2004trust}. Appropriate levels of trust need to be formed among humans and the robotic agent to safely and effectively use robots by a human. That is why trust needs to be modeled, measured, and calibrated during human-robot interaction.

Trust in HRI has a lot in common with trust in HAI, studied at length. Muir et al. \cite{muir1994trust} found the available definitions for trust between humans inconsistent with the nature of HAI based on the multidimensional construct of trust. She defined one of the first trust models for HAI. This model was based on the model of human expectation of automation proposed by Barber et al. \cite{barber1983logic}. Three elements of expectation defined by Barber's model were included in Muir's trust model: the fundamental expectation of persistence, technical competence (skill, rule, and knowledge-based behavior), fiduciary responsibility (intention, power, and authority). Lee and Moray \cite{lee1992trust} built upon Muir's strategy for modeling trust, identifying independent variables that influence trust, and introducing another trust model. Later, other researchers have modeled the operator's trust in automation, considering more factors affecting trust \cite{itoh2000mathematical,farrell2000connectionist,lee2004trust}. These models were finally classified into five groups \cite{moray1999laboratory}: regression-based models \cite{muir1994trust,lee1992trust}, time series models \cite{lee1994trust}, qualitative models \cite{riley1996operator}, argument based probabilistic models\cite{cohen1998trust}, and neural net models \cite{farrell2000connectionist}.

Although there are many similarities between trust in human-robot interaction and trust in human-automation interaction, and there are great similarities among factors affecting trust in automation and human-robot interaction. However, models generated for modeling trust in human-automation interaction are inconsistent with the needs of human-robot interaction. According to Desai et al. \cite{desai2009creating} \say{\textit{these models do not consider some factors that appear while working with robots such as situational awareness, the usability of the interface, physical presence of robots (co-located with human or remote-located), limitations and complexities of the operating environment, workload, task difficulty, etc. which influence HRI considerably}}. Desai et al. \cite{desai2009creating} introduced a schematic of a model considering some factors affecting human-robot trust (e.g., interface usability, user-related factors, and environment-related factors) in conjugate with factors affecting human-automation trust. Latter, Yagoda et al. \cite{yagoda2012you} introduced one of the first models for trust in human-robot interaction. This model was generated based on the different dimensions of a human-robot interaction task and validity assessment of each of these directions by subject matter experts (SMEs) in the human-robot interaction field. Desai et al. \cite{desai2012modeling} also were one of the pioneers in modeling trust in HRI. They generated a more detailed model for trust in human and autonomous robot teleoperation. This model used the Area Under Trust Curve (AUTC) measure to account for an individual's entire interactive experience with the robot. In this model, many factors affecting trust in robots were included (i.e., situation awareness, task difficulty, feedback, long-term interaction, age, and timing of reliability drop). 

The tasking of evolving robotic world is transitioning from strict teleoperation to tasking in which the robots play the role of a team member or a partner in a dyadic or group working. There is a strong correlation between the level of trust in human-robot teammates with the performance of the robotic agent's work, and it also impacts their interaction quality\cite{lee2004trust,desai2012modeling}. According to Xu et al. \cite{xu2016maintaining}, high levels of trust among human-robot teammates often demonstrate great synergy in which matched decision-making capabilities of the human member in the team complements the exhaustive controlling and executing capabilities of the robotic agent. In contrast, a low level of trust among human-robot teammates might cause humans to refuse to delegate tasks to the robotic agent or sometimes decide to disable the robotic agent \cite{xu2016maintaining}. As there is a high correlation between trust and performance of the work in human-robot collaboration, trust can be modeled based on the performance \cite{xu2012trust,sadrfaridpour2016modeling} and modifying the performance of the collaboration based on the human expectations to convince human to show the act of trust toward robotic agent. There are also some trust models based on the performance of robot operation, which is not aimed to modify the performance but also for detecting robotic agents that are not reliable and assigning less critical tasks to them or disregard them while assigning tasks to robotic agents in multi-robot tasks \cite{pippin2014trust}. On the other hand, the human-robot collaboration's performance can also be modeled based on the trust \cite{freedy2007measurement,gao2013modeling}, these models are used to modify the root's trust-related behavior in order to manage and optimize the overall performance of the collaboration.


A prevalent class of human-robot collaborations is supervisory collaboration. There are two roles for an individual in the human-robot teams, supervisor (human) and worker(robot), where the supervisor delegate tasks to the worker and oversee the performance of the operation. Supervisor also has the authority to take control of the robot when the robot is doing something in the wrong way and correct the robot’s mistake. The model of trust for supervisory collaboration, which is presented in \cite{xu2012trust}, is based on the trust in human-human collaboration, generates a quantity showing the compatibility of robot performance with human expectation, and let the robot modify its performance to fulfill human expectations and improve trust. Latter, this trust model was improved \cite{xu2016towards} and more factors affecting trust in supervisory collaboration, such as failure rate in the autonomous agent and the rate of supervisor intervention, were involved in designing the trust model. Online Probabilistic Trust Inference Model (OPTIMo) \cite{xu2015optimo} is another model of trust in human-robot supervisory collaboration which the same research group introduced. This model formulates Bayesian beliefs over human’s trust status in each moment based on the performance of the robot on the task over time to generate an estimate of the real-time human’s trust in the robot.

Modeling trust in human-robot collaboration is mainly used to notify automation whenever the human operator loses trust to let the automation correct its performance. However, human over-reliance needs to be considered in modeling trust and responding by the robot. When trust can be modeled and measured in a real-time manner in human-robot collaboration, it can help the robot repair trust whenever the human starts under-trusting the robot \cite{xu2015optimo}. A real-time model of trust (trust-POMDP) for human-robot peer-to-peer collaboration is introduced in \cite{chen2018planning}, which integrates measured trust in the robot’s decision-making. The trust-POMDP model closes the loop between measured trust by the real-time trust model and robot decision-making process to maximize collaboration performance. This model grants a robot the ability to influence human trust systematically to reduce and increase trust in over-reliance and under-reliance situations, respectively.


Hancock et al. \cite{hancock2011meta} provided a meta-analysis of a significant number of factors affecting trust in human-robot interaction. They evaluate and quantify the effects of human-based, robot-based, and environmental-based factors on trust in human-robot interaction and provide quantitative measurement for the effect of each of these class of factors. They develop a model of human-robot team trust, based on their findings in meta-analysis \cite{hancock2011meta}. In addition to the effect of three classes of factors affecting trust, which they introduced in their meta-analysis, they considered the effects of training implications and designing implications on their final model of trust \cite{sanders2011model}.

Subjective trust measurement techniques are recently being used in companion with objective trust measurement techniques in some of the studies on modeling trust in human-automation interaction and human-robot interaction. Subjective trust measurement techniques are deployed to increase the accuracy and robustness of trust measurements. Chen et al. \cite{chen2018planning} uses machine learning methods to reason for human behavior based on robot behavior and modify robot behavior based on human expectation. There are some studies in human-automation trust, human-computer interaction, and human trust to artificial intelligence that use psycho-physiological measurements for trust modeling \cite{ajenaghughrure2019predictive,gulati2017modelling,gulati2019design}. Khalid et al. \cite{khalid2016exploring} introduced a model for trust modeling in human-robot interaction, which uses facial expressions, voice features, and extracted heat rate features in combination with self-reported trust of human to model trust. They classify human trust in a robot into low, natural, and high trust levels using a Neuro-fuzzy trust classifier.

\section{Trust Model: Inputs and Outputs}

Trust models formulate the effect of factors on the formation and variation of trust in robots. Trust models use factors affecting trust to estimate trust. Since these factors vary in different domains and environments, input factors to the trust models vary based on the application domain. For example, Robinette et al. \cite{robinette2014modeling} models trust in emergency evacuation based on the situational risk (e.g., amount of danger perceived by the human in the environment around him) and agent risk (e.g., agent's behavior and appearance) to model perceived trust by the human and the human's decision to trust the robot's guidance or not. In contrast, \cite{xu2012trust} proposes a trust model for a supervisory collaboration and formulates trust as a function of the robot's success and failure in performing the task. The output of this trust model is closing the loop between human trust and robot function by adjusting the robot's action to improve the collaboration efficiency. Finally, \cite{yagoda2012you} proposes a more general trust model based on team configuration, task, system, context, and team process to scale trust for trust measurement. 

Many of the studies on modeling trust in HRI consider the performance of collaboration as one of the main input elements for their model \cite{xu2015optimo,gao2013modeling,sadrfaridpour2016modeling}. Most of these models consider the effect of performance in conjugate with some other factors. For instance, the OPTIMo probabilistic trust model \cite{xu2015optimo} uses rates of robot's failures and human interventions in conjugate with task performance as inputs to the model to estimate the human's degree of trust in a robotic teammate. Meanwhile, \cite{gao2013modeling} uses the operator's perception of system capabilities, past experience, and training to assess initial trust. Trust gets updated in a loop based on system performance, cognitive workload, and frequency of changes from teleoperation to autonomous operation. This trust model's output is a measure of gain and loss of trust and the impact of these trust changes on collaboration performance. Sadrfaridpour et al. \cite{sadrfaridpour2016modeling} models trust based on human performance (i.e., muscles fatigue and dynamics of recovery), robot performance (i.e., speed of robot doing the specific task), workload, and human expectation of task performance. The output of this model is feedback to the robot to adjust its performance according to operator desires.

\section{Conclusion and future work}

To perform any research on trust in human-robot interaction, we need to define trust clearly. Some of the available definitions of trust that many researchers use are taken from other fields, especially from HAI \cite{lee2004trust}. Recent studies on the nature of trust in human-robot interaction showed that trust has a multidimensional nature \cite{malle2021multidimensional}. Based on this multidimensional nature, any form of trust in robots can be categorized into two categories: performance-based trust, which is mostly based on the factors affecting trust in HAI, and relation-based trust, primarily based on factors affecting trust in human-human interaction. Therefore, we can conclude that trust definitions, which are defined based on the needs of trust in HAI, cannot be inclusive enough to be used in the domain of human-robot trust, and a more inclusive definition for trust, encompassing all aspects of human-robot trust is needed HRI researches.

There are numerous studies on trust violation and trust repair in HRI, most of those study the effects of trust violation and trust repair on a singular moment, and they did not investigate the dynamics of trust loss and trust repair after a specific interaction \cite{lucas2018getting,lee2010receptionist,strohkorb2018ripple,sebo2019don}. However, trust is seen as something that changes over time based on different factors in an interaction. Therefore, the long-term effect of failure and trust repair strategies are understudied. Studies exploring the evolving nature of trust based on the different types of failures and different types of trust repair strategies are missing in the literature. For example, there are no answers to these questions: how does trust loss as the effect of robot failure change with the user's increasing familiarity with a robotic agent? Or how do trust repair strategies help trust repair in long-term interaction where an error has occurred several times and efforts have been made to restore trust several times?

Most of the existing trust models in HRI are developed for a specific form of human-robot interaction, specific task type, or a specific type of robotic agent. For example, in \cite{robinette2014modeling}, a model of trust is specified for evacuation robots; in \cite{desai2012effects} the trust model is specified on robots with shared control, and in \cite{xu2015optimo} the generated model is usable for supervisory collaboration. As present trust models are domain or task specified, they can not be used in other robotic domains, and the result of those can not be compared with each other. Therefore, we can not evaluate any of these trust models. HRI fields lack a general trust model that can be applied to different robotic tasks and domains. Such a model will eliminate the need for creating new trust models for newly emerging tasks.

Trust is a subject of interest for research in many other fields such as psychology, sociology, and even physiology. In these fields, other indicators of trust are used for trust measurement. For example, some studies use physiological indicators, such as oxytocin-related measures \cite{uvnas2005oxytocin,ferreira2018relationship,johnson2011trust}, and objective measures, such as trust games that assess actual investment behavior \cite{chang2010seeing,keri2009sharing}. These methods for trust measurement can be used for measuring trust in human-robot interaction. These trust measurement methods can provide an accurate assessment of trust that does not suffer from the many error probabilities that current trust measurement methods used in human-robot interaction studies suffer from. These measurement methods can also help formulate trust and generate trust models independent of the countless parameters that affect trust.


%

\nocite{}
\bibliography{human-robot trust}

\begin{thebibliography}{100}
\providecommand{\url}[1]{#1}
\csname url@samestyle\endcsname
\providecommand{\newblock}{\relax}
\providecommand{\bibinfo}[2]{#2}
\providecommand{\BIBentrySTDinterwordspacing}{\spaceskip=0pt\relax}
\providecommand{\BIBentryALTinterwordstretchfactor}{4}
\providecommand{\BIBentryALTinterwordspacing}{\spaceskip=\fontdimen2\font plus
\BIBentryALTinterwordstretchfactor\fontdimen3\font minus
  \fontdimen4\font\relax}
\providecommand{\BIBforeignlanguage}[2]{{%
\expandafter\ifx\csname l@#1\endcsname\relax
\typeout{** WARNING: IEEEtran.bst: No hyphenation pattern has been}%
\typeout{** loaded for the language `#1'. Using the pattern for}%
\typeout{** the default language instead.}%
\else
\language=\csname l@#1\endcsname
\fi
#2}}
\providecommand{\BIBdecl}{\relax}
\BIBdecl

\bibitem{lee2004trust}
J.~D. Lee and K.~A. See, ``Trust in automation: Designing for appropriate
  reliance,'' \emph{Human factors}, vol.~46, no.~1, pp. 50--80, 2004.

\bibitem{martelaro2016tell}
N.~Martelaro, V.~C. Nneji, W.~Ju, and P.~Hinds, ``Tell me more designing hri to
  encourage more trust, disclosure, and companionship,'' in \emph{2016 11th
  ACM/IEEE Intl Conf on HRI}.\hskip 1em plus 0.5em minus 0.4em\relax IEEE,
  2016, pp. 181--188.

\bibitem{schaefer2016meta}
K.~E. Schaefer, J.~Y. Chen, J.~L. Szalma, and P.~A. Hancock, ``A meta-analysis
  of factors influencing the development of trust in automation: Implications
  for understanding autonomy in future systems,'' \emph{Human factors},
  vol.~58, no.~3, pp. 377--400, 2016.

\bibitem{robinette2016overtrust}
P.~Robinette, W.~Li, R.~Allen, A.~M. Howard, and A.~R. Wagner, ``Overtrust of
  robots in emergency evacuation scenarios,'' in \emph{2016 11th ACM/IEEE
  International Conference on Human-Robot Interaction (HRI)}.\hskip 1em plus
  0.5em minus 0.4em\relax IEEE, 2016, pp. 101--108.

\bibitem{hancock2011meta}
P.~A. Hancock, D.~R. Billings, K.~E. Schaefer, J.~Y. Chen, E.~J. De~Visser, and
  R.~Parasuraman, ``A meta-analysis of factors affecting trust in human-robot
  interaction,'' \emph{Human factors}, vol.~53, no.~5, pp. 517--527, 2011.

\bibitem{desai2012modeling}
M.~Desai, ``Modeling trust to improve human-robot interaction,'' Ph.D.
  dissertation, University of Massachusetts Lowell, 2012.

\bibitem{sadrfaridpour2016modeling}
B.~Sadrfaridpour, H.~Saeidi, J.~Burke, K.~Madathil, and Y.~Wang, ``Modeling and
  control of trust in human-robot collaborative manufacturing,'' in
  \emph{Robust Intelligence and Trust in Autonomous Systems}.\hskip 1em plus
  0.5em minus 0.4em\relax Springer, 2016, pp. 115--141.

\bibitem{groom2007can}
V.~Groom and C.~Nass, ``Can robots be teammates?: Benchmarks in human--robot
  teams,'' \emph{Interaction Studies}, vol.~8, no.~3, pp. 483--500, 2007.

\bibitem{selkowitz2015effects}
A.~Selkowitz, S.~Lakhmani, J.~Y. Chen, and M.~Boyce, ``The effects of agent
  transparency on human interaction with an autonomous robotic agent,'' in
  \emph{Proceedings of the Human Factors and Ergonomics Society Annual
  Meeting}, vol.~59, no.~1.\hskip 1em plus 0.5em minus 0.4em\relax SAGE
  Publications Sage CA: Los Angeles, CA, 2015, pp. 806--810.

\bibitem{ososky2014determinants}
S.~Ososky, T.~Sanders, F.~Jentsch, P.~Hancock, and J.~Y. Chen, ``Determinants
  of system transparency and its influence on trust in and reliance on unmanned
  robotic systems,'' in \emph{Unmanned Systems Technology XVI}, vol.
  9084.\hskip 1em plus 0.5em minus 0.4em\relax International Society for Optics
  and Photonics, 2014, p. 90840E.

\bibitem{parasuraman1997humans}
R.~Parasuraman and V.~Riley, ``Humans and automation: Use, misuse, disuse,
  abuse,'' \emph{Human factors}, vol.~39, no.~2, pp. 230--253, 1997.

\bibitem{lee2008review}
J.~D. Lee, ``Review of a pivotal human factors article:“humans and
  automation: use, misuse, disuse, abuse”,'' \emph{Human Factors}, vol.~50,
  no.~3, pp. 404--410, 2008.

\bibitem{voth2004new}
D.~Voth, ``A new generation of military robots,'' \emph{IEEE Intelligent
  Systems}, vol.~19, no.~4, pp. 2--3, 2004.

\bibitem{springer2013military}
P.~J. Springer, \emph{Military robots and drones: a reference handbook}.\hskip
  1em plus 0.5em minus 0.4em\relax ABC-CLIO, 2013.

\bibitem{taylor2006perspective}
R.~H. Taylor, ``A perspective on medical robotics,'' \emph{Proceedings of the
  IEEE}, vol.~94, no.~9, pp. 1652--1664, 2006.

\bibitem{taylor2016medical}
R.~H. Taylor, A.~Menciassi, G.~Fichtinger, P.~Fiorini, and P.~Dario, ``Medical
  robotics and computer-integrated surgery,'' in \emph{Springer handbook of
  robotics}.\hskip 1em plus 0.5em minus 0.4em\relax Springer, 2016, pp.
  1657--1684.

\bibitem{yim2003modular}
M.~Yim, K.~Roufas, D.~Duff, Y.~Zhang, C.~Eldershaw, and S.~Homans, ``Modular
  reconfigurable robots in space applications,'' \emph{Autonomous Robots},
  vol.~14, no. 2-3, pp. 225--237, 2003.

\bibitem{moosavian2007free}
S.~A.~A. Moosavian and E.~Papadopoulos, ``Free-flying robots in space: an
  overview of dynamics modeling, planning and control,'' \emph{Robotica},
  vol.~25, no.~5, pp. 537--547, 2007.

\bibitem{ivanov2017robot}
S.~H. Ivanov and C.~Webster, ``The robot as a consumer: a research agenda,'' in
  \emph{Marketing: experience and perspectives” Conference}, 2017, pp.
  29--30.

\bibitem{pandey2016socially}
A.~K. Pandey, ``Socially intelligent robots, the next generation of consumer
  robots and the challenges,'' in \emph{International Conference on ICT
  Innovations}.\hskip 1em plus 0.5em minus 0.4em\relax Springer, 2016, pp.
  41--46.

\bibitem{ososky2013building}
S.~Ososky, D.~Schuster, E.~Phillips, and F.~G. Jentsch, ``Building appropriate
  trust in human-robot teams,'' in \emph{2013 AAAI Spring Symposium Series},
  2013.

\bibitem{atoyan2006trust}
H.~Atoyan, J.-R. Duquet, and J.-M. Robert, ``Trust in new decision aid
  systems,'' in \emph{Proceedings of the 18th Conference on l'Interaction
  Homme-Machine}, 2006, pp. 115--122.

\bibitem{wagner2011recognizing}
A.~R. Wagner and R.~C. Arkin, ``Recognizing situations that demand trust,'' in
  \emph{2011 RO-MAN}.\hskip 1em plus 0.5em minus 0.4em\relax IEEE, 2011, pp.
  7--14.

\bibitem{freedy2007measurement}
A.~Freedy, E.~DeVisser, G.~Weltman, and N.~Coeyman, ``Measurement of trust in
  human-robot collaboration,'' in \emph{2007 International Symposium on
  Collaborative Technologies and Systems}.\hskip 1em plus 0.5em minus
  0.4em\relax IEEE, 2007, pp. 106--114.

\bibitem{cameron2015framing}
D.~Cameron, J.~Aitken, E.~Collins, L.~Boorman, A.~Chua, S.~Fernando, O.~McAree,
  U.~Martinez~Hernandez, and J.~Law, ``Framing factors: the importance of
  context and the individual in understanding trust in human-robot
  interaction,'' 2015.

\bibitem{oleson2011antecedents}
K.~E. Oleson, D.~R. Billings, V.~Kocsis, J.~Y. Chen, and P.~A. Hancock,
  ``Antecedents of trust in human-robot collaborations,'' in \emph{2011 IEEE
  International Multi-Disciplinary Conference on Cognitive Methods in Situation
  Awareness and Decision Support (CogSIMA)}.\hskip 1em plus 0.5em minus
  0.4em\relax IEEE, 2011, pp. 175--178.

\bibitem{xu2015optimo}
A.~Xu and G.~Dudek, ``Optimo: Online probabilistic trust inference model for
  asymmetric human-robot collaborations,'' in \emph{2015 10th ACM/IEEE
  International Conference on Human-Robot Interaction (HRI)}.\hskip 1em plus
  0.5em minus 0.4em\relax IEEE, 2015, pp. 221--228.

\bibitem{frennert2017case}
S.~Frennert, H.~Eftring, and B.~{\"O}stlund, ``Case report: Implications of
  doing research on socially assistive robots in real homes,''
  \emph{International Journal of Social Robotics}, vol.~9, no.~3, pp. 401--415,
  2017.

\bibitem{ahtinen2020learning}
A.~Ahtinen and K.~Kaipainen, ``Learning and teaching experiences with a
  persuasive social robot in primary school--findings and implications from a
  4-month field study,'' in \emph{International Conference on Persuasive
  Technology}.\hskip 1em plus 0.5em minus 0.4em\relax Springer, 2020, pp.
  73--84.

\bibitem{reeder2010breakbot}
S.~Reeder, L.~Kelly, B.~Kechavarzi, and S.~Sabanovic, ``Breakbot: a social
  motivator for the workplace,'' in \emph{Proceedings of the 8th ACM Conference
  on Designing Interactive Systems}, 2010, pp. 61--64.

\bibitem{wolbring2014social}
G.~Wolbring and S.~Yumakulov, ``Social robots: views of staff of a disability
  service organization,'' \emph{International journal of social robotics},
  vol.~6, no.~3, pp. 457--468, 2014.

\bibitem{malle2021multidimensional}
B.~F. Malle and D.~Ullman, ``A multidimensional conception and measure of
  human-robot trust,'' in \emph{Trust in Human-Robot Interaction}.\hskip 1em
  plus 0.5em minus 0.4em\relax Elsevier, 2021, pp. 3--25.

\bibitem{park2020multifaceted}
S.~Park, ``Multifaceted trust in tourism service robots,'' \emph{Annals of
  Tourism Research}, vol.~81, p. 102888, 2020.

\bibitem{hoff2015trust}
K.~A. Hoff and M.~Bashir, ``Trust in automation: Integrating empirical evidence
  on factors that influence trust,'' \emph{Human factors}, vol.~57, no.~3, pp.
  407--434, 2015.

\bibitem{desai2012effects}
M.~Desai, M.~Medvedev, M.~V{\'a}zquez, S.~McSheehy, S.~Gadea-Omelchenko,
  C.~Bruggeman, A.~Steinfeld, and H.~Yanco, ``Effects of changing reliability
  on trust of robot systems,'' in \emph{2012 7th ACM/IEEE International
  Conference on Human-Robot Interaction (HRI)}.\hskip 1em plus 0.5em minus
  0.4em\relax IEEE, 2012, pp. 73--80.

\bibitem{lucas2018getting}
G.~M. Lucas, J.~Boberg, D.~Traum, R.~Artstein, J.~Gratch, A.~Gainer,
  E.~Johnson, A.~Leuski, and M.~Nakano, ``Getting to know each other: The role
  of social dialogue in recovery from errors in social robots,'' in
  \emph{Proceedings of the 2018 acm/ieee international conference on
  human-robot interaction}, 2018, pp. 344--351.

\bibitem{salem2015would}
M.~Salem, G.~Lakatos, F.~Amirabdollahian, and K.~Dautenhahn, ``Would you trust
  a (faulty) robot? effects of error, task type and personality on human-robot
  cooperation and trust,'' in \emph{2015 10th ACM/IEEE International Conference
  on Human-Robot Interaction (HRI)}.\hskip 1em plus 0.5em minus 0.4em\relax
  IEEE, 2015, pp. 1--8.

\bibitem{natarajan2020effects}
M.~Natarajan and M.~Gombolay, ``Effects of anthropomorphism and accountability
  on trust in human robot interaction,'' in \emph{Proceedings of the 2020
  ACM/IEEE International Conference on Human-Robot Interaction}, 2020, pp.
  33--42.

\bibitem{boyce2015effects}
M.~W. Boyce, J.~Y. Chen, A.~R. Selkowitz, and S.~G. Lakhmani, ``Effects of
  agent transparency on operator trust,'' in \emph{Proceedings of the Tenth
  Annual ACM/IEEE International Conference on Human-Robot Interaction Extended
  Abstracts}, 2015, pp. 179--180.

\bibitem{yang2017evaluating}
X.~J. Yang, V.~V. Unhelkar, K.~Li, and J.~A. Shah, ``Evaluating effects of user
  experience and system transparency on trust in automation,'' in \emph{2017
  12th ACM/IEEE International Conference on Human-Robot Interaction
  (HRI}.\hskip 1em plus 0.5em minus 0.4em\relax IEEE, 2017, pp. 408--416.

\bibitem{lazanyi2017dispositional}
K.~Lazanyi and G.~Maraczi, ``Dispositional trust—do we trust autonomous
  cars?'' in \emph{2017 IEEE 15th International Symposium on Intelligent
  Systems and Informatics (SISY)}.\hskip 1em plus 0.5em minus 0.4em\relax IEEE,
  2017, pp. 000\,135--000\,140.

\bibitem{tay2014stereotypes}
B.~Tay, Y.~Jung, and T.~Park, ``When stereotypes meet robots: the double-edge
  sword of robot gender and personality in human--robot interaction,''
  \emph{Computers in Human Behavior}, vol.~38, pp. 75--84, 2014.

\bibitem{robinette2013building}
P.~Robinette, A.~R. Wagner, and A.~M. Howard, ``Building and maintaining trust
  between humans and guidance robots in an emergency,'' in \emph{2013 AAAI
  Spring Symposium Series}, 2013.

\bibitem{you2018human}
S.~You and L.~P. Robert~Jr, ``Human-robot similarity and willingness to work
  with a robotic co-worker,'' in \emph{Proceedings of the 2018 ACM/IEEE
  International Conference on Human-Robot Interaction}, 2018, pp. 251--260.

\bibitem{mumm2011human}
J.~Mumm and B.~Mutlu, ``Human-robot proxemics: physical and psychological
  distancing in human-robot interaction,'' in \emph{Proceedings of the 6th
  international conference on Human-robot interaction}, 2011, pp. 331--338.

\bibitem{obaid2016stop}
M.~Obaid, E.~B. Sandoval, J.~Z{\l}otowski, E.~Moltchanova, C.~A. Basedow, and
  C.~Bartneck, ``Stop! that is close enough. how body postures influence
  human-robot proximity,'' in \emph{2016 25th IEEE International Symposium on
  Robot and Human Interactive Communication (RO-MAN)}.\hskip 1em plus 0.5em
  minus 0.4em\relax IEEE, 2016, pp. 354--361.

\bibitem{walters2011long}
M.~L. Walters, M.~A. Oskoei, D.~S. Syrdal, and K.~Dautenhahn, ``A long-term
  human-robot proxemic study,'' in \emph{2011 RO-MAN}.\hskip 1em plus 0.5em
  minus 0.4em\relax IEEE, 2011, pp. 137--142.

\bibitem{wang2016trust}
N.~Wang, D.~V. Pynadath, and S.~G. Hill, ``Trust calibration within a
  human-robot team: Comparing automatically generated explanations,'' in
  \emph{2016 11th ACM/IEEE International Conference on Human-Robot Interaction
  (HRI)}.\hskip 1em plus 0.5em minus 0.4em\relax IEEE, 2016, pp. 109--116.

\bibitem{strohkorb2018ripple}
S.~Strohkorb~Sebo, M.~Traeger, M.~Jung, and B.~Scassellati, ``The ripple
  effects of vulnerability: The effects of a robot's vulnerable behavior on
  trust in human-robot teams,'' in \emph{Proceedings of the 2018 ACM/IEEE
  International Conference on Human-Robot Interaction}, 2018, pp. 178--186.

\bibitem{sebo2019don}
S.~S. Sebo, P.~Krishnamurthi, and B.~Scassellati, ``“i don't believe you”:
  Investigating the effects of robot trust violation and repair,'' in
  \emph{2019 14th ACM/IEEE International Conference on Human-Robot Interaction
  (HRI)}.\hskip 1em plus 0.5em minus 0.4em\relax IEEE, 2019, pp. 57--65.

\bibitem{bainbridge2011benefits}
W.~A. Bainbridge, J.~W. Hart, E.~S. Kim, and B.~Scassellati, ``The benefits of
  interactions with physically present robots over video-displayed agents,''
  \emph{International Journal of Social Robotics}, vol.~3, no.~1, pp. 41--52,
  2011.

\bibitem{desai2013impact}
M.~Desai, P.~Kaniarasu, M.~Medvedev, A.~Steinfeld, and H.~Yanco, ``Impact of
  robot failures and feedback on real-time trust,'' in \emph{2013 8th ACM/IEEE
  International Conference on Human-Robot Interaction (HRI)}.\hskip 1em plus
  0.5em minus 0.4em\relax IEEE, 2013, pp. 251--258.

\bibitem{goodrich2001experiments}
M.~A. Goodrich, D.~R. Olsen, J.~W. Crandall, and T.~J. Palmer, ``Experiments in
  adjustable autonomy,'' in \emph{Proceedings of IJCAI Workshop on autonomy,
  delegation and control: interacting with intelligent agents}.\hskip 1em plus
  0.5em minus 0.4em\relax Seattle, WA, 2001, pp. 1624--1629.

\bibitem{rau2009effects}
P.~P. Rau, Y.~Li, and D.~Li, ``Effects of communication style and culture on
  ability to accept recommendations from robots,'' \emph{Computers in Human
  Behavior}, vol.~25, no.~2, pp. 587--595, 2009.

\bibitem{li2010cross}
D.~Li, P.~P. Rau, and Y.~Li, ``A cross-cultural study: Effect of robot
  appearance and task,'' \emph{International Journal of Social Robotics},
  vol.~2, no.~2, pp. 175--186, 2010.

\bibitem{m2011measuring}
K.~M~Tsui, M.~Desai, H.~A~Yanco, H.~Cramer, and N.~Kemper, ``Measuring
  attitudes towards telepresence robots,'' \emph{Intl journal of intelligent
  control and systems}, vol.~16, 2011.

\bibitem{flook2019impact}
R.~Flook, A.~Shrinah, L.~Wijnen, K.~Eder, C.~Melhuish, and S.~Lemaignan, ``On
  the impact of different types of errors on trust in human-robot interaction:
  Are laboratory-based hri experiments trustworthy?'' \emph{Interaction
  Studies}, vol.~20, no.~3, pp. 455--486, 2019.

\bibitem{baker2018toward}
A.~L. Baker, E.~K. Phillips, D.~Ullman, and J.~R. Keebler, ``Toward an
  understanding of trust repair in human-robot interaction: Current research
  and future directions,'' \emph{ACM Transactions on Interactive Intelligent
  Systems (TiiS)}, vol.~8, no.~4, pp. 1--30, 2018.

\bibitem{schaefer2016measuring}
K.~E. Schaefer, ``Measuring trust in human robot interactions: Development of
  the “trust perception scale-hri”,'' in \emph{Robust Intelligence and
  Trust in Autonomous Systems}.\hskip 1em plus 0.5em minus 0.4em\relax
  Springer, 2016, pp. 191--218.

\bibitem{jian2000foundations}
J.-Y. Jian, A.~M. Bisantz, and C.~G. Drury, ``Foundations for an empirically
  determined scale of trust in automated systems,'' \emph{International journal
  of cognitive ergonomics}, vol.~4, no.~1, pp. 53--71, 2000.

\bibitem{yagoda2012you}
R.~E. Yagoda and D.~J. Gillan, ``You want me to trust a robot? the development
  of a human--robot interaction trust scale,'' \emph{International Journal of
  Social Robotics}, vol.~4, no.~3, pp. 235--248, 2012.

\bibitem{madsen2000measuring}
M.~Madsen and S.~Gregor, ``Measuring human-computer trust,'' in \emph{11th
  australasian conference on information systems}, vol.~53.\hskip 1em plus
  0.5em minus 0.4em\relax Citeseer, 2000, pp. 6--8.

\bibitem{cannon2011team}
J.~A. Cannon-Bowers and C.~Bowers, ``Team development and functioning.'' 2011.

\bibitem{law2021trust}
T.~Law and M.~Scheutz, ``Trust: Recent concepts and evaluations in human-robot
  interaction,'' in \emph{Trust in Human-Robot Interaction}.\hskip 1em plus
  0.5em minus 0.4em\relax Elsevier, 2021, pp. 27--57.

\bibitem{pedersen2018simulations}
B.~K. M.~K. Pedersen, K.~E. Andersen, S.~K{\"o}slich, B.~C. Weigelin, and
  K.~Kuusinen, ``Simulations and self-driving cars: A study of trust and
  consequences,'' in \emph{Companion of the 2018 ACM/IEEE International
  Conference on Human-Robot Interaction}, 2018, pp. 205--206.

\bibitem{chen2018planning}
M.~Chen, S.~Nikolaidis, H.~Soh, D.~Hsu, and S.~Srinivasa, ``Planning with trust
  for human-robot collaboration,'' in \emph{Proceedings of the 2018 ACM/IEEE
  International Conference on Human-Robot Interaction}, 2018, pp. 307--315.

\bibitem{xie2019robot}
Y.~Xie, I.~P. Bodala, D.~C. Ong, D.~Hsu, and H.~Soh, ``Robot capability and
  intention in trust-based decisions across tasks,'' in \emph{2019 14th
  ACM/IEEE International Conference on Human-Robot Interaction (HRI)}.\hskip
  1em plus 0.5em minus 0.4em\relax IEEE, 2019, pp. 39--47.

\bibitem{rossi2018impact}
A.~Rossi, K.~Dautenhahn, K.~L. Koay, and M.~L. Walters, ``The impact of
  peoples’ personal dispositions and personalities on their trust of robots
  in an emergency scenario,'' \emph{Paladyn, Journal of Behavioral Robotics},
  vol.~9, no.~1, pp. 137--154, 2018.

\bibitem{jayaraman2018trust}
S.~K. Jayaraman, C.~Creech, L.~P. Robert~Jr, D.~M. Tilbury, X.~J. Yang, A.~K.
  Pradhan, and K.~M. Tsui, ``Trust in av: An uncertainty reduction model of
  av-pedestrian interactions,'' in \emph{Companion of the 2018 ACM/IEEE
  International Conference on Human-Robot Interaction}, 2018, pp. 133--134.

\bibitem{weigelin2018trust}
B.~C. Weigelin, M.~Mathiesen, C.~Nielsen, K.~Fischer, and J.~Nielsen, ``Trust
  in medical human-robot interactions based on kinesthetic guidance,'' in
  \emph{2018 27th IEEE International Symposium on Robot and Human Interactive
  Communication (RO-MAN)}.\hskip 1em plus 0.5em minus 0.4em\relax IEEE, 2018,
  pp. 901--908.

\bibitem{christensen2019reducing}
A.~B. Christensen, C.~R. Dam, C.~Rasle, J.~E. Bauer, R.~A. Mohamed, and L.~C.
  Jensen, ``Reducing overtrust in failing robotic systems,'' in \emph{2019 14th
  ACM/IEEE International Conference on Human-Robot Interaction (HRI)}.\hskip
  1em plus 0.5em minus 0.4em\relax IEEE, 2019, pp. 542--543.

\bibitem{gombolay2018robotic}
M.~Gombolay, X.~J. Yang, B.~Hayes, N.~Seo, Z.~Liu, S.~Wadhwania, T.~Yu,
  N.~Shah, T.~Golen, and J.~Shah, ``Robotic assistance in the coordination of
  patient care,'' \emph{The International Journal of Robotics Research},
  vol.~37, no.~10, pp. 1300--1316, 2018.

\bibitem{robinette2015timing}
P.~Robinette, A.~M. Howard, and A.~R. Wagner, ``Timing is key for robot trust
  repair,'' in \emph{Intl Conf on social robotics}.\hskip 1em plus 0.5em minus
  0.4em\relax Springer, 2015, pp. 574--583.

\bibitem{rossi2017timing}
A.~Rossi, K.~Dautenhahn, K.~L. Koay, and M.~L. Walters, ``How the timing and
  magnitude of robot errors influence peoples’ trust of robots in an
  emergency scenario,'' in \emph{International Conference on Social
  Robotics}.\hskip 1em plus 0.5em minus 0.4em\relax Springer, 2017, pp. 42--52.

\bibitem{lee2010receptionist}
M.~K. Lee, S.~Kiesler, and J.~Forlizzi, ``Receptionist or information kiosk:
  how do people talk with a robot?'' in \emph{Proceedings of the 2010 ACM
  conference on Computer supported cooperative work}, 2010, pp. 31--40.

\bibitem{hamacher2016believing}
A.~Hamacher, N.~Bianchi-Berthouze, A.~G. Pipe, and K.~Eder, ``Believing in
  bert: Using expressive communication to enhance trust and counteract
  operational error in physical human-robot interaction,'' in \emph{2016 25th
  IEEE international symposium on robot and human interactive communication
  (RO-MAN)}.\hskip 1em plus 0.5em minus 0.4em\relax IEEE, 2016, pp. 493--500.

\bibitem{tolmeijer2020taxonomy}
S.~Tolmeijer, A.~Weiss, M.~Hanheide, F.~Lindner, T.~M. Powers, C.~Dixon, and
  M.~L. Tielman, ``Taxonomy of trust-relevant failures and mitigation
  strategies,'' in \emph{Proceedings of the 2020 ACM/IEEE International
  Conference on Human-Robot Interaction}, 2020, pp. 3--12.

\bibitem{muir1994trust}
B.~M. Muir, ``Trust in automation: Part i. theoretical issues in the study of
  trust and human intervention in automated systems,'' \emph{Ergonomics},
  vol.~37, no.~11, pp. 1905--1922, 1994.

\bibitem{barber1983logic}
B.~Barber, ``The logic and limits of trust,'' 1983.

\bibitem{lee1992trust}
J.~Lee and N.~Moray, ``Trust, control strategies and allocation of function in
  human-machine systems,'' \emph{Ergonomics}, vol.~35, no.~10, pp. 1243--1270,
  1992.

\bibitem{itoh2000mathematical}
M.~Itoh and K.~Tanaka, ``Mathematical modeling of trust in automation: Trust,
  distrust, and mistrust,'' in \emph{Proceedings of the human factors and
  ergonomics society annual meeting}, vol.~44, no.~1.\hskip 1em plus 0.5em
  minus 0.4em\relax SAGE Publications Sage CA: Los Angeles, CA, 2000, pp.
  9--12.

\bibitem{farrell2000connectionist}
S.~Farrell and S.~Lewandowsky, ``A connectionist model of complacency and
  adaptive recovery under automation.'' \emph{Journal of Experimental
  Psychology: Learning, Memory, and Cognition}, vol.~26, no.~2, p. 395, 2000.

\bibitem{moray1999laboratory}
N.~Moray and T.~Inagaki, ``Laboratory studies of trust between humans and
  machines in automated systems,'' \emph{Transactions of the Institute of
  Measurement and Control}, vol.~21, no. 4-5, pp. 203--211, 1999.

\bibitem{lee1994trust}
J.~D. Lee and N.~Moray, ``Trust, self-confidence, and operators' adaptation to
  automation,'' \emph{Intl journal of human-computer studies}, vol.~40, no.~1,
  pp. 153--184, 1994.

\bibitem{riley1996operator}
V.~Riley, ``Operator reliance on automation: Theory and data,''
  \emph{Automation and human performance: Theory and applications}, pp. 19--35,
  1996.

\bibitem{cohen1998trust}
M.~S. Cohen, R.~Parasuraman, and J.~T. Freeman, ``Trust in decision aids: A
  model and its training implications,'' in \emph{in Proc. Command and Control
  Research and Technology Symp}.\hskip 1em plus 0.5em minus 0.4em\relax
  Citeseer, 1998.

\bibitem{desai2009creating}
M.~Desai, K.~Stubbs, A.~Steinfeld, and H.~Yanco, ``Creating trustworthy robots:
  Lessons and inspirations from automated systems,'' 2009.

\bibitem{xu2016maintaining}
A.~Xu and G.~Dudek, ``Maintaining efficient collaboration with trust-seeking
  robots,'' in \emph{2016 IEEE/RSJ International Conference on Intelligent
  Robots and Systems (IROS)}.\hskip 1em plus 0.5em minus 0.4em\relax IEEE,
  2016, pp. 3312--3319.

\bibitem{xu2012trust}
{Xu, Anqi and Dudek, Gregory}, ``Trust-driven interactive visual navigation for
  autonomous robots,'' in \emph{2012 IEEE International Conference on Robotics
  and Automation}.\hskip 1em plus 0.5em minus 0.4em\relax IEEE, 2012, pp.
  3922--3929.

\bibitem{pippin2014trust}
C.~Pippin and H.~Christensen, ``Trust modeling in multi-robot patrolling,'' in
  \emph{2014 IEEE International Conference on Robotics and Automation
  (ICRA)}.\hskip 1em plus 0.5em minus 0.4em\relax IEEE, 2014, pp. 59--66.

\bibitem{gao2013modeling}
F.~Gao, A.~Clare, J.~Macbeth, and M.~Cummings, ``Modeling the impact of
  operator trust on performance in multiple robot control,'' in \emph{2013 AAAI
  Spring Symposium Series}, 2013.

\bibitem{xu2016towards}
A.~Xu and G.~Dudek, ``Towards modeling real-time trust in asymmetric
  human--robot collaborations,'' in \emph{Robotics Research}.\hskip 1em plus
  0.5em minus 0.4em\relax Springer, 2016, pp. 113--129.

\bibitem{sanders2011model}
T.~Sanders, K.~E. Oleson, D.~R. Billings, J.~Y. Chen, and P.~A. Hancock, ``A
  model of human-robot trust: Theoretical model development,'' in
  \emph{Proceedings of the human factors and ergonomics society annual
  meeting}, vol.~55, no.~1.\hskip 1em plus 0.5em minus 0.4em\relax SAGE
  Publications Sage CA: Los Angeles, CA, 2011, pp. 1432--1436.

\bibitem{ajenaghughrure2019predictive}
I.~B. Ajenaghughrure, S.~C. Sousa, I.~J. Kosunen, and D.~Lamas, ``Predictive
  model to assess user trust: a psycho-physiological approach,'' in
  \emph{Proceedings of the 10th Indian Conference on Human-Computer
  Interaction}, 2019, pp. 1--10.

\bibitem{gulati2017modelling}
S.~Gulati, S.~Sousa, and D.~Lamas, ``Modelling trust: An empirical
  assessment,'' in \emph{IFIP Conference on Human-Computer Interaction}.\hskip
  1em plus 0.5em minus 0.4em\relax Springer, 2017, pp. 40--61.

\bibitem{gulati2019design}
{Gulati, Siddharth and Sousa, Sonia and Lamas, David}, ``Design, development
  and evaluation of a human-computer trust scale,'' \emph{Behaviour \&
  Information Technology}, vol.~38, no.~10, pp. 1004--1015, 2019.

\bibitem{khalid2016exploring}
H.~M. Khalid, L.~W. Shiung, P.~Nooralishahi, Z.~Rasool, M.~G. Helander, L.~C.
  Kiong, and C.~Ai-vyrn, ``Exploring psycho-physiological correlates to trust:
  Implications for human-robot-human interaction,'' in \emph{Proceedings of the
  human factors and ergonomics society annual meeting}, vol.~60, no.~1.\hskip
  1em plus 0.5em minus 0.4em\relax SAGE Publications Sage CA: Los Angeles, CA,
  2016, pp. 697--701.

\bibitem{robinette2014modeling}
P.~Robinette, A.~R. Wagner, and A.~M. Howard, ``Modeling human-robot trust in
  emergencies,'' in \emph{2014 AAAI Spring Symposium Series}, 2014.

\bibitem{uvnas2005oxytocin}
K.~Uvnas-Moberg and M.~Petersson, ``Oxytocin, a mediator of anti-stress,
  well-being, social interaction, growth and healing,'' \emph{Z Psychosom Med
  Psychother}, vol.~51, no.~1, pp. 57--80, 2005.

\bibitem{ferreira2018relationship}
F.~Ferreira, R.~Lopes~da Costa, L.~Pereira, C.~Jer{\'o}nimo, and {\'A}.~Dias,
  ``The relationship between chemical of happiness, chemical of stress,
  leadership, motivation and organizational trust: A case study on brazilian
  workers,'' \emph{The relationship between chemical of happiness, chemical of
  stress, leadership, motivation and organizational trust: a case study on
  Brazilian workers}, no.~2, pp. 89--100, 2018.

\bibitem{johnson2011trust}
N.~D. Johnson and A.~A. Mislin, ``Trust games: A meta-analysis,'' \emph{Journal
  of Economic Psychology}, vol.~32, no.~5, pp. 865--889, 2011.

\bibitem{chang2010seeing}
L.~J. Chang, B.~B. Doll, M.~van’t Wout, M.~J. Frank, and A.~G. Sanfey,
  ``Seeing is believing: Trustworthiness as a dynamic belief,'' \emph{Cognitive
  psychology}, vol.~61, no.~2, pp. 87--105, 2010.

\bibitem{keri2009sharing}
S.~Keri, I.~Kiss, and O.~Kelemen, ``Sharing secrets: oxytocin and trust in
  schizophrenia,'' \emph{Social neuroscience}, vol.~4, no.~4, pp. 287--293,
  2009.

\end{thebibliography}
\bibliographystyle{IEEEtran}

\end{document}